%% file: icipv7.tex
\documentclass{article}
\usepackage{spconf,amsmath,epsfig}

\usepackage{amstext,amssymb,amsbsy,amsmath,amscd}
\usepackage{graphics,graphicx,epsfig,subfigure,enumerate,color,pstricks,cancel,rotating,theorem,array,longtable,multirow,multicol}
\usepackage{algorithmic,algorithm}

\input{prelim.tex}

\newtheorem{theorem}{Theorem}
\newtheorem{corollaire}{Corollary}

\title{Region-based active contour with noise and shape priors}
\name{F. Lecellier$^a$, S. Jehan-Besson$^a$, J. Fadili$^a$, G. Aubert$^b$, M. Revenu$^a$, E.Saloux$^c$}
\address{$^a$GREYC UMR 6072 Caen France, $^b$Laboratoire J.A. Dieudonn\'e Nice France, $^c$CHU Caen France}
\begin{document}
%\ninept
%
\maketitle
\begin{abstract}
In this paper, we propose to combine formally noise and shape priors in region-based active contours. On the one hand, we use the general framework of exponential family as a prior model for noise. On the  other hand, translation and scale invariant Legendre moments are considered to incorporate the shape prior (e.g. fidelity to a reference shape). The combination of the two prior terms in the active contour functional yields the final evolution equation whose evolution speed is rigorously derived using shape derivative tools. Experimental results on both synthetic images and real life cardiac echography data clearly demonstrate the robustness to initialization and noise, flexibility and large potential applicability of our segmentation algorithm.
\end{abstract}


\vspace{-0.5em}
\section{Introduction}
\label{sec:intro}

The current work is devoted to the segmentation of regions of a priori known shape in noisy images using region-based active contours \cite{Cremers02, Tsai03,gastaud03}. This method
%Region-based active contours 
allows the use of photometric image properties, such as texture and noise, as well as geometric properties such as the shape of the object to be segmented. The shape prior can prove very useful in cases where the  object is occluded or partially missing. Furthermore, by including an a priori on the shape, sensitivity of the active contour model to initialization will also be alleviated. 

On the one hand, attempts to incorporate shape priors have been proposed by some authors using various methods, such as diffusion snakes \cite{Cremers02} or distance function \cite{Tsai03,gastaud03,Leventon00}.
On the other hand, there are only few proposals in the literature that tried to take benefit of a noise prior \cite{refregier04,Lecellier_icassp06,Galland05} within region-based active contours. In these works, image features (e.g. intensity) are considered as random variables whose distribution belongs to some parametric family which is chosen according to the physical acquisition model of the considered images. However, to the best of our knowledge, shape and noise priors have never been combined, at least formally, in active contour models. This would enable to perform the segmentation of poor noisy images, beyond the simple classical white Gaussian noise model, with a strong shape constraint. Example of such data is encountered in echocardiographic images. For instance, in echographic data, it is notably well known that under appropriate conditions (large number of randomly located scatters), the Rayleigh distribution is well suited to model the noise \cite{Goodman76}.

The main contribution of this paper is to combine formally noise and shape priors in region-based active contours for segmentation purposes. In order to fix ideas, let us consider a region of interest $\Omega$ in the image. We propose to find the partition of the image that minimizes the following generic criterion which is able to handle both noise and shape priors:
\vspace{-0.3em}
\begin{eqnarray}
J(\Omega)=  \nonumber 
\int_{\Omega} f_n(\x,\Omega) d\x+\alpha \: d(\Omega,\Omega_{ref}) 
%+ \beta \: E_b(\Gamma)
\end{eqnarray}
where $\Omega_{ref}$ represents the reference region shape, and $\x=[x,y]^T$ stands for the location of the pixel.
%and $\Gamma$ to the boundary of $\Omega$. The energy term $E_b$ is a regularization term balanced with a positive real parameter $\beta$.

The first term corresponds to the noise prior term. This term takes benefit of statistical properties of the image intensity. It is based on functions of parametric probability density functions (pdf) belonging to the exponential family. Probability models with these common features include Normal, Bernoulli, Binomial, Poisson, Gamma, Beta, Rayleigh, etc. These models are the most commonly encountered in imaging acquisition systems. This term is detailed in Section \ref{sec:noise}.
The second term $d(\Omega,\Omega_{ref})$ corresponds to the shape prior. Shapes are here described using scale and translation invariant Legendre moments as in \cite{Foulonneau03}. With such a shape descriptor, the registration step is avoided. This term is discussed in Section \ref{sec:shape}.

The evolution equation of the deformable curve is deduced from the functional to minimize using shape derivative tools \cite{Zolesio} and the framework set in \cite{Jehan03, Aubert03}. 

This paper is organized as follows: we briefly remind the shape derivation tools in Section \ref{sec:tools}. The noise model term is presented in Section \ref{sec:noise}. In Section \ref{sec:shape}, we introduce the shape prior model and the invariances that were added. The segmentation algorithm is presented in Section \ref{sec:algo}. Experimental results are discussed in Section \ref{sec:results}. We finally conclude and give some perspectives.

\section{Shape derivative tools}
\label{sec:tools}
In order to be comprehensive, we here give a brief summary of the shape derivation theory. The interested reader may refer to \cite{Zolesio, Jehan03, Aubert03} for further details.

Let $\U$ be a class of domains (open, regular bounded sets, i.e. $C^2$) of $\R^n$, and
$\Rg$ an element of $\U$. The boundary $\partial\Rg$ of $\Rg$ is sometimes
denoted by $\C$.

The region-based term is expressed as a domain integral of a function $f$ named descriptor of the region :
\vspace{-0.3em}
\begin{eqnarray}
\label{eq:terme_gal_base_region}
{J_r}(\Rg)=\int_{\Rg} f(\x,\Rg) d\x
\end{eqnarray}
In the general case, this descriptor may depend on the domain such as the descriptors introduced thereafter for noise and shape priors.
The derivation of this term is performed using domain derivation tools. We apply a fundamental theorem \cite{Zolesio} which establishes a relation between the Eulerian derivative of $J_r(\Rg)$ in the direction $\V$, and the domain derivative of $f$ denoted $f_s(\x,\Rg,\V)$:
\vspace{-0.5em}
\begin{eqnarray}
<J_r'(\Rg),\V>&=&\int_{\Rg} f_s(\x,\Rg,\V)d\x \nonumber \\ & & - \int_{\partial\Rg} f(\x,\Rg) \, (\V \cdot \N) d{\mathbf{a}}(\x)
\end{eqnarray}
where $\N$ is the unit inward normal to $\partial\Omega$, $d{\mathbf{a}}$ its area element. The first integral comes from the dependence of the descriptor $f(\x,\Rg)$ upon the region while the second term comes from the evolution of the region itself.

From the shape derivative, we can deduce the evolution equation that will drive the active contour towards a minimum of the criterion.
%The boundary energy term $E_b$ may be chosen as the curve length and derived classically using calculus of variation or shape derivation tools \cite{Zolesio}.

Let us suppose that the shape derivative of the region $\Om$ may be written as follows:
\vspace{-0.5em}
\begin{equation}
\label{eq:Eul_der}
<J_{r}'(\Omega),\V>=-\int_{\partial \Rg} \mathbf{v}(\x,\Om) ( \V(\x) \cdot \N(\x))  d\mathbf{a}(\x) 
\end{equation}
We can then deduce the following evolution equation: 
%(more details may be found in \cite{Jehan03}) from the minimization of criterion \ref{eq:terme_gal_base_region} with 2 regions (inside and outside $\Omega$) :
\vspace{-0.5em}
\begin{eqnarray*}
%\frac{\partial\Gamma(p,\tau)}{\partial\tau}=\left(\mathbf{v}_{in}(\x,\Om_{in})- \mathbf{v}_{out}(\x,\Om_{out})+\beta \kappa(\x)\right)\N(\x)
\frac{\partial\Gamma(p,\tau)}{\partial\tau}=\mathbf{v}(\x,\Om)\,\N(\x)
\end{eqnarray*}
with $\Gamma(\tau=0)={\Gamma}_0$, $\x=\Gamma(p,\tau)$. 

\section{The noise model}
\label{sec:noise}
In this section we focus our attention on the noise model. The chosen descriptor for this part is:
\vspace{-0.5em}
\begin{eqnarray}
f_n(\x,\Omega)=\Phi( p(\y(\x),\veta)
\end{eqnarray}
where $p$ is the pdf of some image features $\y(\x)\in \R^d$ whose associated parameters are denoted by $\veta$, and $\Phi$ is at least $C^1$.

In our study, we consider that $p$ belongs to the exponential family. This family is comprehensive enough to cover noise models in most image acquisition systems encountered in practice, e.g. Gaussian, Exponential, Poisson, Rayleigh to cite a few. The multi-parameter exponential families are naturally indexed by a $k$-dimensional real {\em{natural parameter}} vector $\veta=(\eta_1,\ldots,\eta_k)^T$ and a $k$-dimensional {\em{natural sufficient statistic}} vector $\vT=(T_1,\ldots,T_k)^T$. A simple example is the normal family when both the location and the scale parameters are unknown ($k=2$). Formally, the pdf of a vector of random variables $\Y$ belonging to the $k$-parameter {\em canonical} exponential family is:
\vspace{-0.5em}
\begin{equation}
\label{eqdef3}
p(\y,\veta) = h(\y) \exp [\langle\veta,\vT(\y)\rangle - A(\veta)]
\end{equation}
where $\langle\veta,\vT\rangle$ denotes the scalar product.

This statistical criterion is now derived according to the domain in order to deduce the evolution equation of the active contour. %Relative proofs are given in\cite{Lecellier_icassp06}. 
For the sake of simplicity, we denote $\veta$ for the natural parameter of a pdf of the exponential family and its finite sample estimate over the domain (without a slight abuse of notation, this should be $\hat{\veta}$). 
\begin{theorem}
\label{th:gene}
The G\^ateaux derivative, in the direction of $\V$, of the functional $J_{n}(\Om)=\int_{\Om} \Phi(p(\y(\x),\veta(\Om)))d{\mathbf{a}}(\x)$ where $p(.)$ belongs to the multi-parameter exponential family with natural hyperparameter vector $\veta$, is:
\vspace{-0.5em}
\begin{eqnarray}
\label{derE}
<J_{n}'(\Om),\V>= -\int_{\partial\Omega}{\Phi(p(\y))(\V \cdot \N) d{\mathbf{a}}(\x)} \nonumber
\\ + \int_{\Om} p(\y) \Phi'(p(\y))\langle\nabla_{\V}\veta,\T(\y)-\nabla A(\veta)\rangle d\x
\end{eqnarray}
with $\nabla_{\V}\veta$ the G\^ateaux derivative of $\veta$ in the direction of $\V$, and $\langle \x,\y \rangle$ the scalar product of vectors $\x$ and $\y$.
\end{theorem}

In a finite sample setting, when using the ML estimator, we can replace $\nabla A(\veta)$ by $\overline{\T(\Y)}$ (the 1st order sample moment of $\T(\Y))$. Thus, when using the -log-likelihood function, the second term becomes equal to $\int_{\Om}\langle\nabla_{\V}\veta,\T(\y)-\overline{\T(\Y)})\rangle d\x$, and hence vanishes. The following corollary follows:

\begin{corollaire}
\label{th:der_multipar}
The G\^ateaux derivative, in the direction of $\V$, of the functional $J_{n}(\Om)=-\int_{\Om} \log(p(\y(\x),\mathbf{\veta_{_{ML}}}(\Om))d{\mathbf{a}}(\x)$ when $\mathbf{\vetaML}$ is the ML estimate, is the following: 
\vspace{-0.5em}
\[
<J_{n}'(\Om),\V>=\int_{\partial{\Om}}(\log{(p(\y(\x),\mathbf{\veta_{_{ML}}}(\Om)))}( \V \cdot\N)  d{\mathbf{a}}(\x)
\]
\end{corollaire}

These general results can be easily specialized to some pdf of interest (e.g. Gaussian, Rayleigh, etc). We let the reader refer to \cite{Lecellier_icassp06} for more details.%The interested reader may refer to \cite{Lecellier_icassp06} for detailed expressions of the evolution speed and equation for the case of Gaussian, Rayleigh, Poisson and Exponential distributions.

\section{The shape prior model}
\label{sec:shape}

The shape prior is used as an additional fidelity term (e.g. to a reference shape), designed to make the behaviour of the segmentation algorithm more robust to occlusion and missing data and to alleviate initialization issues. Here, orthogonal Legendre moments with scale and translation invariance were used as shape descriptors \cite{Foulonneau03}. Indeed, moments \cite{teh88} give a region-based compact representation of shapes through the projection of their characteristic functions on an orthogonal basis such as Legendre polynomials.

The shape prior is then defined as the Euclidean distance between the moments of the evolving region and ones of the reference shape,
\vspace{-0.5em}
\begin{equation}
d(\Omega,\Omega_{ref})=\|\lambda(\Omega)-\lambda(\Omega_{ref})\|_2^2
\end{equation}
where $\lambda(\Om)$ are the moments of the region $\Omega$.
%\subsection{Moments in image analysis} 
In practice, infinite moment expansion is generally limited to a sufficient finite number resulting in a good approximation of the original shape. The criterion then reduces to:
\vspace{-0.5em}
\begin{equation}
\label{eq:shape_prior_n}
d(\Omega,\Omega_{ref})=\sum_{p,q}^{p+q \leq N}(\lambda_{pq}(\Omega)-\lambda_{pq}(\Omega_{ref}))^2
\end{equation}
where the $\lambda_{pq}$ are defined as follows, using the geometric moments  $M_{pq}$ and the coefficients $a_{pq}$ of the Legendre polynomials \cite{teh88}:
\vspace{-0.5em}
\begin{equation}
\lambda_{pq}=C_{pq}\sum_{u=0}^p\sum_{v=0}^q a_{pu} a_{qv} M_{uv}
\end{equation}
where $C_{pq}=\frac{(2p+1)(2q+1)}{4}$, $M_{pq}(\Omega)=\int_{\Omega}x^p y^q dxdy$, and the Legendre polynomials are defined as :
\begin{equation*}
P_p(x)=\sum_{k=0}^{p}a_{pk}x^k=\frac{1}{2^p p!}\frac{d^p}{dx^p}(x^2-1)^p.
\end{equation*}

%\subsection{Scale and translation invariance}
In general, the reference shape can have different orientation and size compared to the shape to be segmented. This will then necessitate an explicit registration step in order to realign the two shapes. In order to avoid this generally problematic registration step, we here use scale and translation invariant Legendre moments as in \cite{Foulonneau03}. 
%Firstly scale invariants, and secondly translation, we describe now scale invariants. 
%To deal with scale invariance, we supposed that the shape occupy one quarter of the image. 
In the geometric moments definition, the scale invariance is embodied as a normalization term: $\frac{1}{\Omega^{(p+q+2)/2}}$. As far as translation invariance is concerned, we replace $x$ and $y$ in the geometric moments $M_{pq}$ by $x-\bar x$ and $y-\bar y$, $(\bar{x},\bar{y})$ are the shape barycenter coordinates.

%We have now to compute $\bar x$ and $\bar y$ which are respectively the x-center and y-center of the shape.
%$\bar x=\frac{M_{01}}{M_{00}},\bar y=\frac{M_{10}}{M_{00}}$

%We have now a shape which is normalized and centered. 

The derivation of the criterion (\ref{eq:shape_prior_n}) is relatively complex. We here give the main formula.
\vspace{-0.5em}
\begin{eqnarray*}
<d'(\Omega,\Omega_{ref}),\V>=\sum_{u,v}^{u+v \leq N}A_{uv}\left(H_{uv}+L_{uv}\right)\N
\end{eqnarray*}
where
\begin{small}
$$A_{uv}=2\sum_{p,q}^{p+q \leq N}(\lambda_{pq}-\lambda^{ref}_{pq})C_{pq}a_{pu}a_{qv} \quad H_{uv}=\frac{(x-\bar x)^u(y-\bar y)^v}{\Omega^{(u+v+2)/2}}$$
$$L_{uv}=\frac{u\bar x M_{u-1,v}}{\Omega^{3/2}}(1-x)+\frac{v\bar y M_{u,v-1}}{\Omega^{3/2}}(1-y) -\frac{(u+v+2)M_{u,v}}{2\Omega}$$
\end{small}
The reader may refer to \cite{Foulonneau03} for further details.

\section{Segmentation with joint noise and shape priors}
\label{sec:algo}

The region-based active contour functional to be minimized is finally written as:
\vspace{-0.5em}
\begin{small}
\begin{eqnarray}
\label{eq:crit:tot}
J(\Omega_{in},\Omega_{out})&=&\int_{\Omega_{in}} f_n(\x,\Omega_{in})\,d\x +\alpha \, d(\Omega_{in},\Omega_{ref})\nonumber\\&+& \int_{\Omega_{out}} f_n(\x,\Omega_{out})\,d\x + \beta \, E_b(\Gamma)
\end{eqnarray}
\end{small}
where we assign a specific noise model to the background (outside) region, possibly different from the noise model of the object (inside) region. The energy term $E_b$ is a regularization term balanced with a positive real parameter $\beta$. It can be chosen as the curve length and classically derived using calculus of variation or shape derivation tools.

To drive this functional towards its minimum, the geometrical PDE corresponding to (\ref{eq:crit:tot}) is iteratively run without the shape prior, then the shape prior term is updated, and the active contour evolves again by running the PDE with the shape prior. This procedure is repeated until convergence. This iterative optimization scheme has a flavour of coordinate relaxation. At this stage, it is worth pointing out some major differences between our algorithm and the one developed in \cite{Foulonneau03}. The first one is that we here consider both photometric (noise) and geometrical (shape) priors, while \cite{Foulonneau03} focused on the shape prior and did not considered noisy data. This difference has a clear impact on the evolution algorithm since those authors propose to run the evolution equation only once without the shape prior and then incorporate the shape prior in the evolution. This is fundamentally different from our alternating scheme.

\begin{algorithm}
\begin{algorithmic}[1]
\REPEAT 
\STATE Evolution using noise prior for $n$ iterations
\REPEAT 
\STATE Evolution using shape prior for 1 iteration.
\UNTIL Maximum shape speed $<$ threshold
\UNTIL Convergence
\end{algorithmic}
\caption{Evolution algorithm of the active contour}
\label{algo:evol}
\end{algorithm}
\vspace{-0.5em}
\section{Experimental results}
\label{sec:results}
\begin{figure}
\center
\begin{tabular} {c c c}
\includegraphics[width=0.27\linewidth]{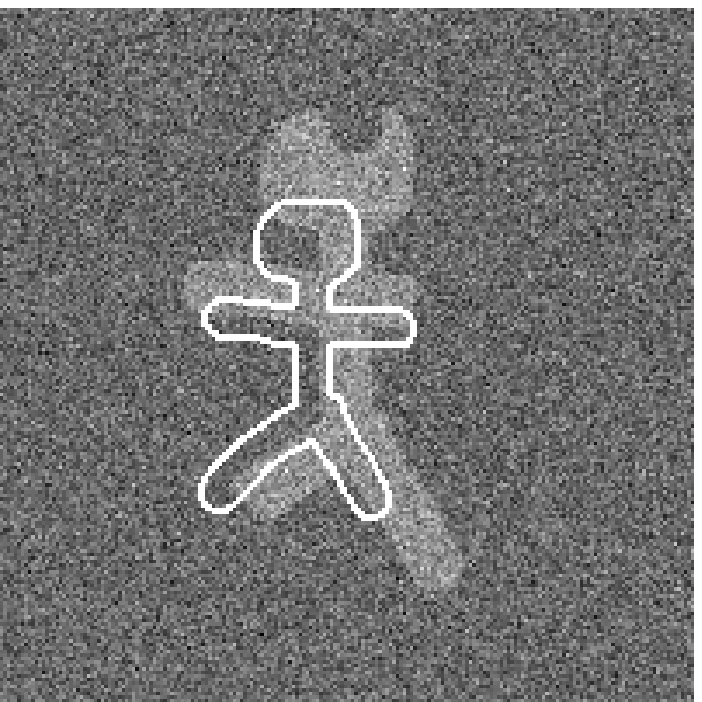} &
\includegraphics[width=0.27\linewidth]{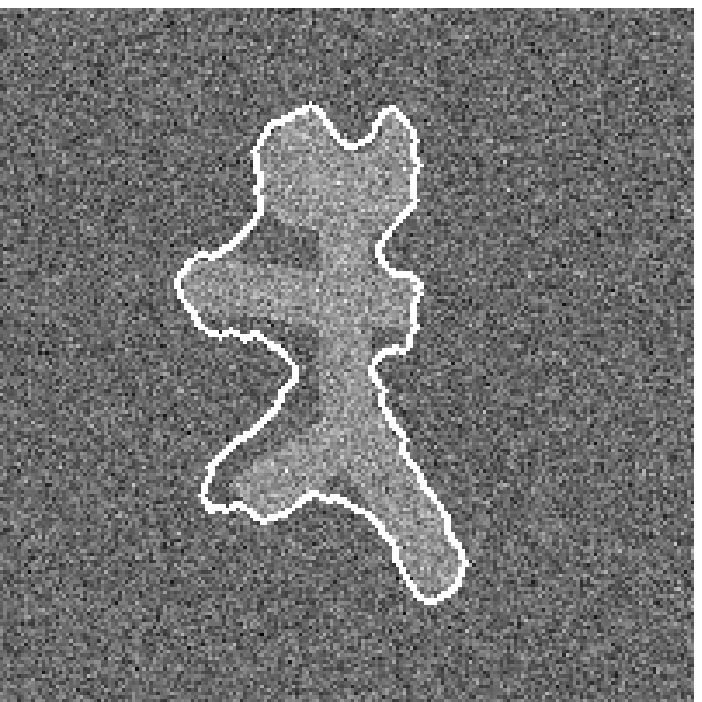} &
\includegraphics[width=0.27\linewidth]{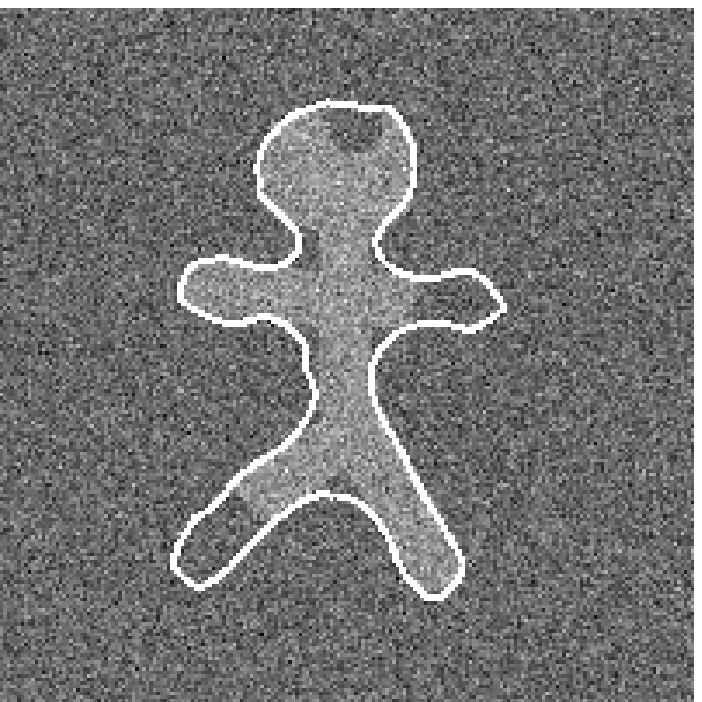} \\
(a) & (b) & (c)
\end{tabular}
\caption{\small a. Noisy image with initial contour, b. Final contour without shape prior, c. Final contour with shape prior.}
\label{fig:bon}
\end{figure}

The above evolution scheme was applied on some synthetic data with $\y(\x)=I(\x)$ the image intensity. Fig.\ref{fig:bon}.(a) depicts a shape corrupted by an additive white Gaussian noise with SNR=1, with the initial curve. To bring to the fore the contribution of the shape prior term, parts of the objects are deliberately missing. Fig.\ref{fig:bon}.(b) (resp. (c)) shows the segmentation result with the noise model (Gaussian) but without (resp. with) the shape prior. As expected, one can clearly see that: (i) without a shape prior, the final curve sticks to the apparent boundaries of object, (ii) owing to the shape prior, the algorithm managed to recover properly the missing parts of the object. Furthermore, in addition to its robustness to missing data, we have also observed that the shape prior allows to mitigate initialization issues. As far as the noise prior is concerned, choosing the appropriate model has a clear impact on the quality of the results as it has been shown in \cite{Lecellier_icassp06}.

\begin{figure}
\center
\begin{tabular}{c c}
\includegraphics[height=0.25\linewidth]{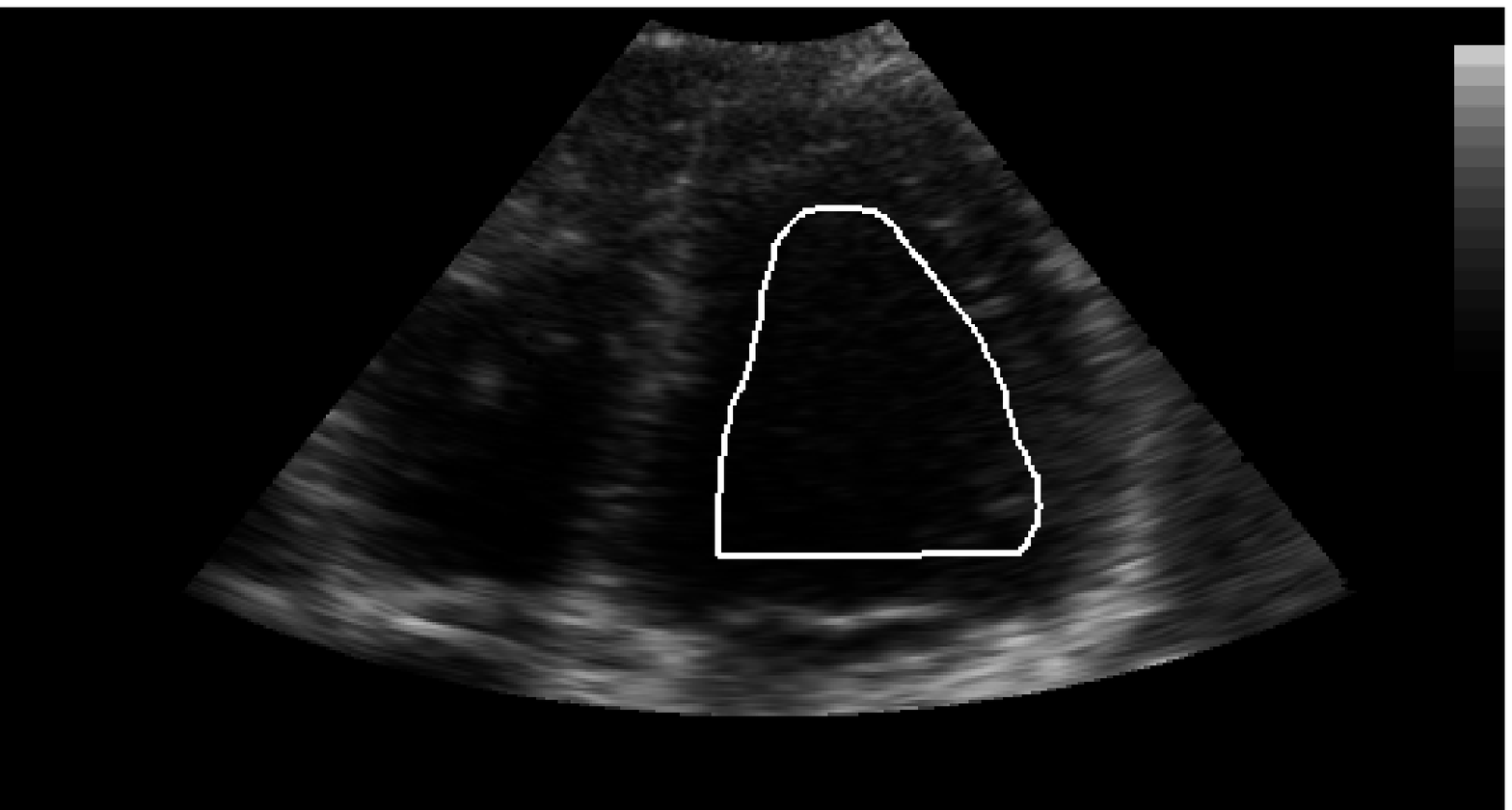} &
\includegraphics[height=0.25\linewidth]{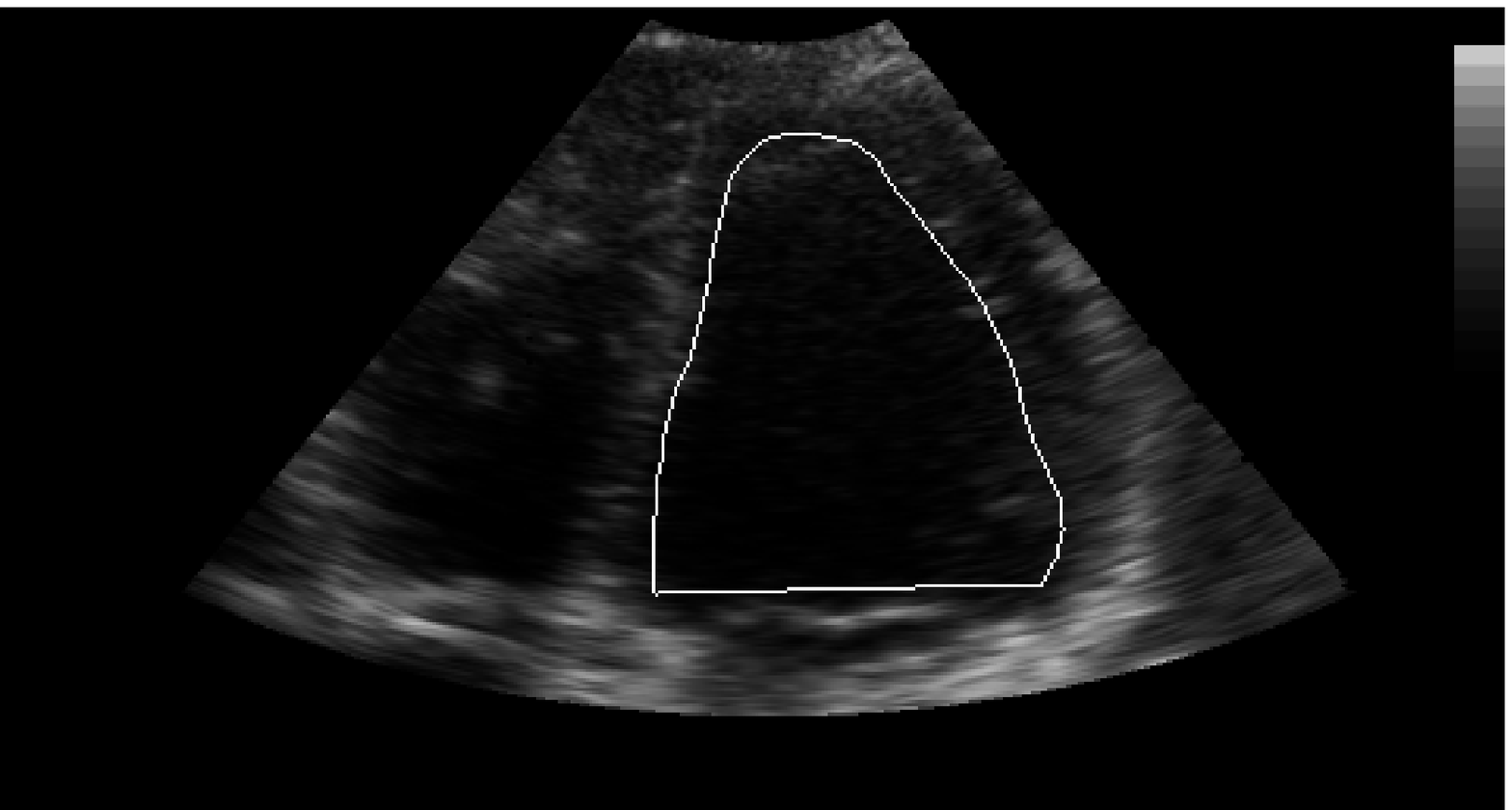} \\
(a)& (b)\\
\includegraphics[height=0.25\linewidth]{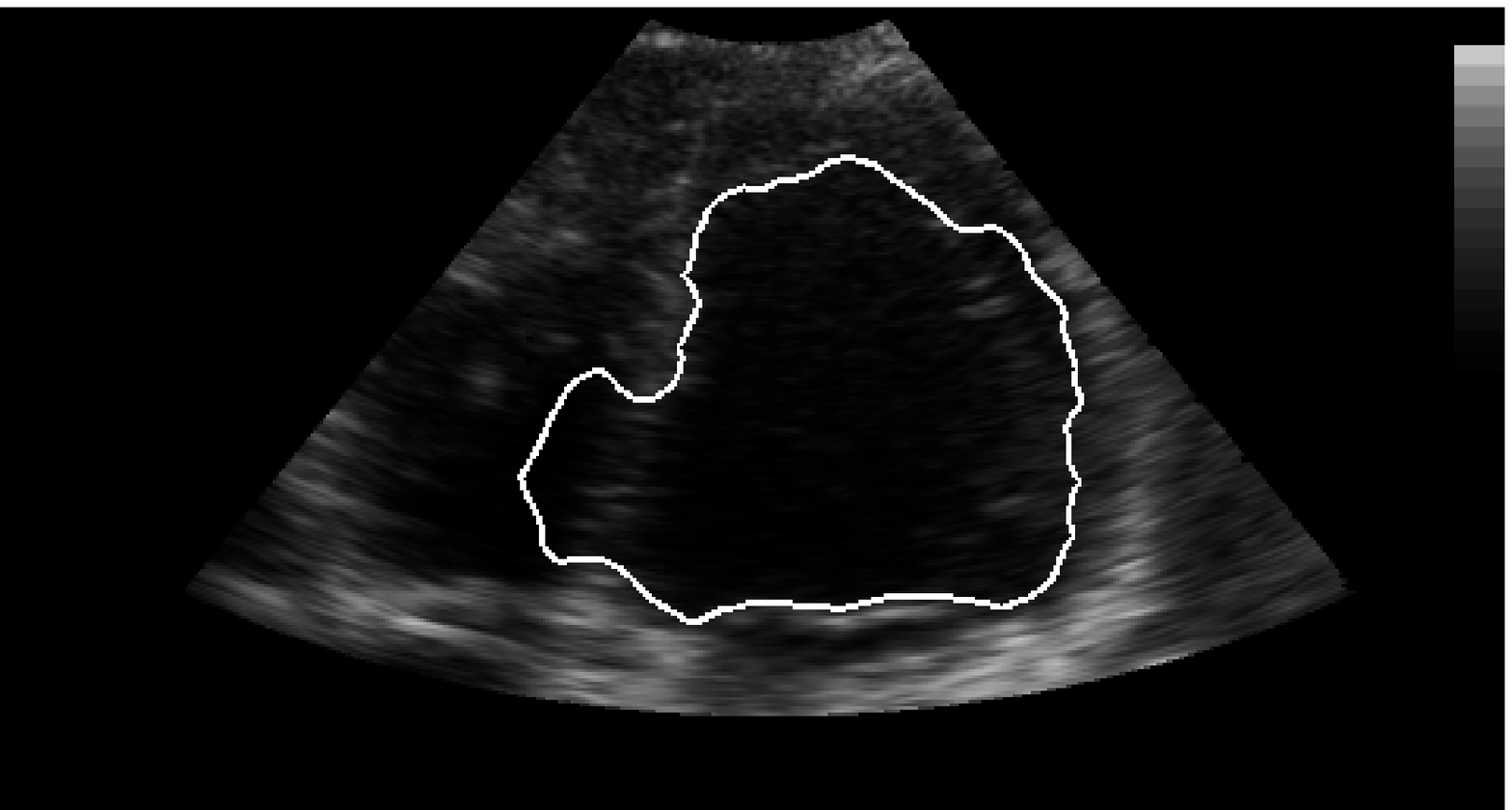} &
\includegraphics[height=0.25\linewidth]{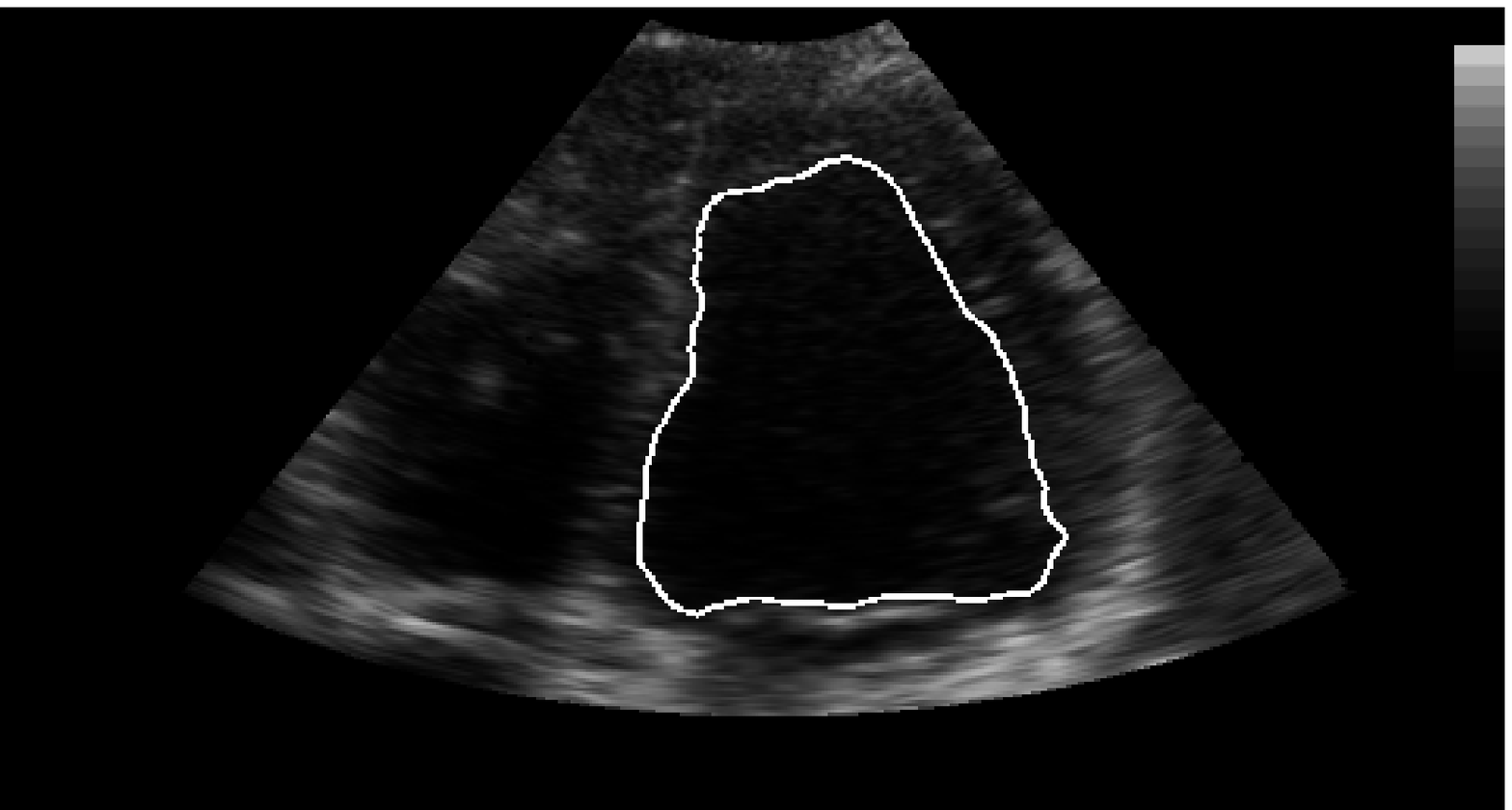} \\
(c) & (d) \\
\includegraphics[height=0.25\linewidth]{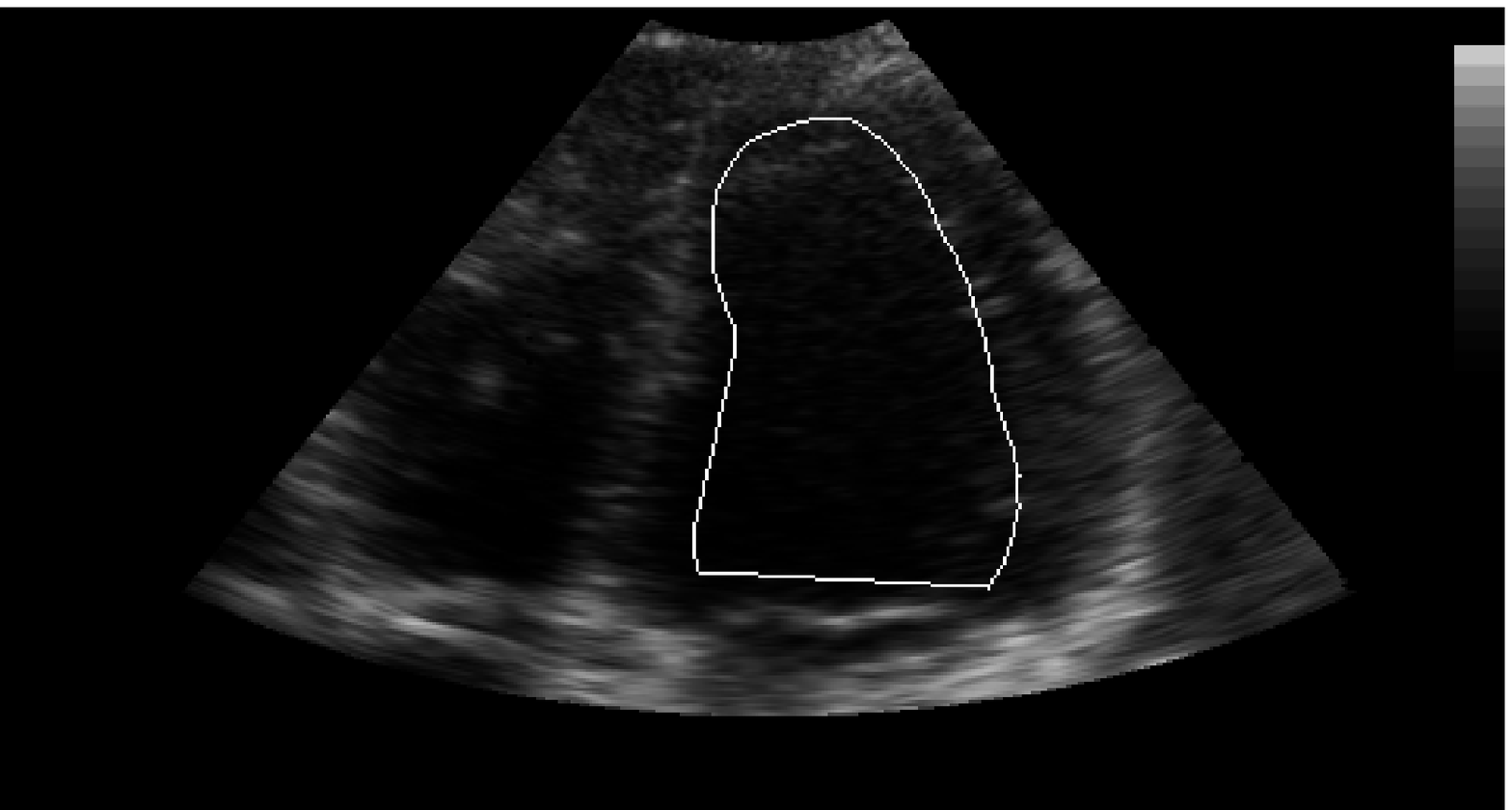} &
\includegraphics[width=4cm,height=2cm]{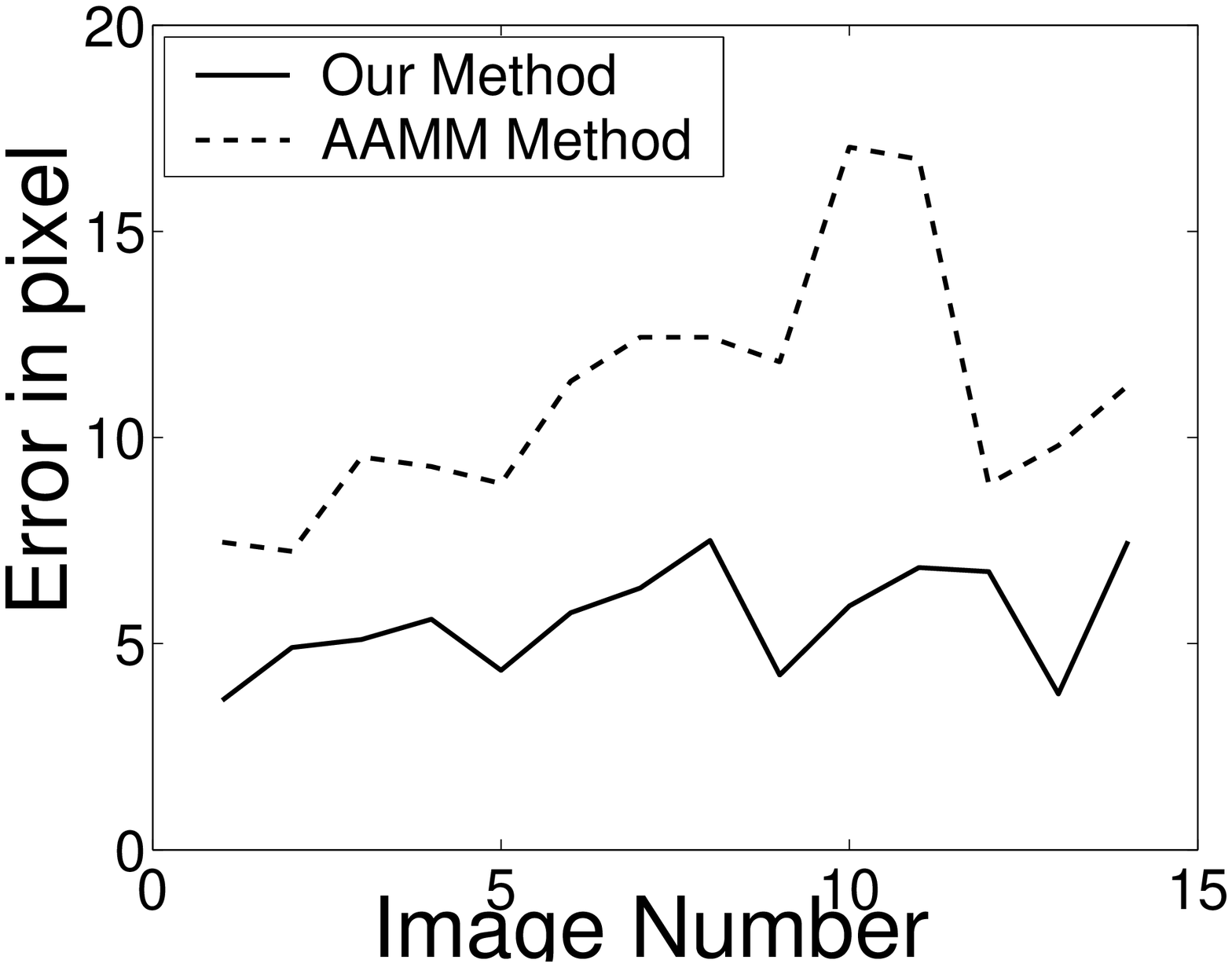}\\
(e) & (f)
\end{tabular}
\caption{\small a. Echocardiographic image with initial contour, b. Contour draw by an expert, c. Final contour without shape prior, d. Final contour with shape prior, e. Final contour using AAMM method, f. Hamming distance for one echocardiographic sequence of 14 images.}
\label{fig:echo}
\end{figure}

We then tested our approach on real echocardiographic images. As the Rayleigh distribution is well suited to model the noise in these data \cite{Goodman76}, this noise model was used in Corollary \ref{th:der_multipar}. The original image (Fig.\ref{fig:echo}.(a)) is shown with the initial contour position. We compared the result of our method (fig.\ref{fig:echo}), with (d) and without (c) the shape prior, to an expert manual segmentation (b), and a segmentation provided by the Active Appearance and Motion Model (AAMM) method (e) designed for echocardiography \cite{Stegmann00b,Lebosse05}. Again, the saliency of our method is obvious. Our method gives the closest segmentation to the expert manual delineation. This is quantitavely by the Hamming distance plots (f), showing that our method outperformes AAMM.
%For a more efficient comparison, we have also compute the Euclidian distance between expert segmentation and our segmentation. The mean distance is 6 pixels with a standard deviation of 1. Those results are very promising, and with a real learning, we think that they will be improve.
\vspace{-0.5em}
\vspace{-0.5em}

\section{Conclusion and perspectives}
\label{sec:conclu}
This paper concerns the incorporation of both noise and shape priors in region-based active contours. The evolution of the active contour is derived from a global criterion that combines statistical image properties and geometrical information. Statistical image properties take benefit of a prespecified noise model defined using parametric pdfs belonging to the exponential family. The geometrical information consists in minimizing the distance between Legendre moments of the shape and those of a reference. The Legendre moments are designed to be scale and translation invariant in order to avoid the registration step. The combination of these terms gives accurate results on both synthetic noisy images and real echocardiographic data. Our ongoing research is now directed towards the integration of a complete shape learning step.
\small
\bibliographystyle{IEEEbib}
\bibliography{biblio}

\end{document}

%% file: prelim.tex
\newcommand{\veta}{{\boldsymbol{\eta}}}

\newcommand{\vT}{{\mathbf{T}}}

\newcommand{\x}{\mathbf{x}}
\newcommand{\y}{\mathbf{y}}
\newcommand{\Y}{\mathbf{Y}}

\newcommand{\V}{\mathbf{V}}
\newcommand{\U}{\mathcal{U}}
\newcommand{\R}{\mathbb{R}}
\newcommand{\T}{\mathbf{T}}
\newcommand{\N}{\mathbf{N}}

\newcommand{\C}{\Gamma}

\newcommand{\Rg}{\Omega}

\newcommand{\Om}{\Omega}

\newcommand{\be}{\begin{eqnarray}}
\newcommand{\ee}{\end{eqnarray}}

\newcommand{\vetaML}{\hat{\mathbf{\eta}}_{ML}}